\documentclass{article}

    \usepackage[final,nonatbib]{tccml_neurips_2020}

\usepackage[utf8]{inputenc} % allow utf-8 input
\usepackage[T1]{fontenc}    % use 8-bit T1 fonts
\usepackage{hyperref}       % hyperlinks
\usepackage{url}            % simple URL typesetting
\usepackage{booktabs}       % professional-quality tables
\usepackage{amsfonts}       % blackboard math symbols
\usepackage{nicefrac}       % compact symbols for 1/2, etc.
\usepackage{microtype}      % microtypography
\usepackage{cite}
\usepackage[]{todonotes}
\usepackage{adjustbox}
\usepackage{tikz}
\usepackage{gensymb}
\usetikzlibrary{shapes,arrows}

\tikzstyle{decision} = [diamond, draw, fill=blue!20, 
    text width=4.5em, text badly centered, node distance=3cm, inner sep=0pt]
\tikzstyle{block} = [rectangle, draw, fill=blue!20, 
    text width=2em, text badly centered, rounded corners, minimum height=2em]
\tikzstyle{line} = [draw, -latex']
\tikzstyle{cloud} = [draw, ellipse,fill=red!20, node distance=3cm, minimum height=2em]

\def\BibTeX{{\rm B\kern-.05em{\sc I\kern-.025em b}\kern-.08em
    T\kern-.1667em\lower.7ex\hbox{E}\kern-.125emX}}

\title{VConstruct: Filling Gaps in Chl-a Data Using a Variational Autoencoder}

\author{%
  Matthew Ehrler \\
  Department of Computer Science\\
  University of Victoria\\
  Victoria B.C \\
  \texttt{mehrler@uvic.ca}\\
  \and
  \textbf{Neil Ernst} \\
  Department of Computer Science\\
  University of Victoria\\
  Victoria B.C \\
  \texttt{nernst@uvic.ca} \\
}

\begin{document}

\maketitle

\begin{abstract}
  Remote sensing of Chlorophyll-a is vital in monitoring climate change. Chlorphyll-a measurements give us an idea of the algae concentrations in the ocean, which lets us monitor ocean health. However, a common problem is that the satellites used to gather the data are commonly obstructed by clouds and other artifacts. This means that time series data from satellites can suffer from spatial data loss. 
  There are a number of algorithms that are able to reconstruct the missing parts of these images to varying degrees of accuracy, with Data INterpolating Empirical Orthogonal Functions (DINEOF) being the current standard. However, DINEOF is slow, suffers from accuracy loss in temporally homogenous waters, reliant on temporal data, and only able to generate a single potential reconstruction.
  We propose a machine learning approach to reconstruction of Chlorophyll-a data using a Variational Autoencoder (VAE). Our accuracy results to date are competitive with but slightly less accurate than DINEOF. We show the benefits of our method including vastly decreased computation time and ability to generate multiple potential reconstructions. Lastly, we outline our planned improvements and future work. 
\end{abstract}

\section{Introduction}

Phytoplankton and ocean colour are considered ``Essential Climate Variables" for measuring and predicting climate systems and ocean health \cite{Bojinski2014}. Measuring phytoplankton and ocean colour is cost effective on a global scale as well as relevant to climate models. Chlorophyll-a (Chl-a) is a commonly used
metric to estimate phytoplankton levels (measured in units of ${mg}/m^3$) and can be derived from ocean colour \cite{Alvain2005}. Additionally, Chl-a can also be used to detect harmful algae blooms which can be fatal to marine life \cite{Sirjacobs2011}. As climate change progresses harmful algae blooms will increase in frequency. An increase of 2\degree C in sea temperature will double the window of opprtunity for Harmful Algae Blooms in the Puget Sound\cite{Moore2008}. Several different satellites provide Chl-a measurements but the Sentinel-3 mission\footnote{https://sentinel.esa.int/web/sentinel/missions/sentinel-3} will be the focus of this paper. 

One of the biggest problems faced when using these measurements is the loss of spatial data due to clouds, sunglint or various other factors which can affect the atmospheric correction process \cite{Sirjacobs2011}. Various algorithms exist to reconstruct the missing data with the most effective being those based off of extracting Empirical Orthographic Functions (EOF)
from the data \cite{Taylor2013}. The most accurate and commonly used of these algorithms is Data INterpolating Empirical Orthogonal Functions (DINEOF) which iteratively calculate EOFs based on the input data \cite{Hilborn2018,Taylor2013}. DINEOF is fairly slow \cite{Taylor2013} and performs poorly in more temporally homogenous waters such as a river mouth \cite{Hilborn2018}.

Machine learning has also been successful in reconstructing Chl-a data. Park et al. use a tree based model to reconstruct algae in polar regions \cite{Park2019}. This method is effective but requires knowledge of the domain to properly tune it. This makes it much less generalizable and therefore less effective than DINEOF as DINEOF works with no a priori knowledge. 
DINCAE is a very new machine learning approach to reconstructing data \cite{Barth2020}, which has also been shown to work on Chl-a \cite{Han2020}. DINCAE is accurate, but shares a drawback with DINEOF in that it can only generate a single possible reconstruction. Being able to see multiple potential reconstructions and potentially select a better one based on data that may not be able to be easily or quickly incorporated into the model.
For example if we had Chl-a concentrations manually measured from missing areas, we could then generate reconstructions until we find one that better matches the measured values. This would be much faster than changing DINCAE or DINEOF to incorporate the new data.

The approach we outline in this paper is based on the Variational Autoencoder (VAE) from Kingma et al. \cite{Kingma2014} as well as Attribute2Image's improvements in making generated images less random \cite{Yan2016}. The dimensionality reduction in a VAE is somewhat similar to the Singular Value Decomposition (SVD) used in DINEOF. The potential to leverage performance improvements using quantum annealing machines with VAEs was another motivation. \cite{Khoshaman2018}. 

In this paper we apply a model similar to Attribute2Image as well as Ivanov et al's inpainting model to Chl-a data from the Salish Sea area surrounding Vancouver Island \cite{Yan2016,Ivanov2018}. We compare it to the industry standard DINEOF using experiments modeled after Hilborn et al.'s experiments \cite{Hilborn2018}. This area was chosen as it contains both areas of
high and low temporal homogeneity in terms of algae concentrations, which was determined by Hilborn et al to be something DINEOF is sensitive to \cite{Hilborn2018}.

\section{Method}
\subsection{Dataset and Preprocessing}
The dataset we use comes from the Algae Explorer Project\footnote{https://algaeexplorer.ca/}. This project used 1566 images taken daily from 2016-04-25 to 2020-09-30. For our experiments we use a 250x250 pixel slice from each day to create a 1566x250x250 dataset.

We then preprocess the data for DINEOF using a similar process to Hilborn et al. \cite{Hilborn2018}. The data for VConstruct uses a similar process to Han et al. \cite{Han2020}. 

We select five days for testing at random from all days that have very low cloud cover. This allows us to add artificial clouds and measure accuracy by comparing to the original complete image.

\subsection{DINEOF Testing}
As we are using different satellite data than Hilborn et al., we cannot compare directly to their results and need to devise a similar experiment \cite{Hilborn2018}. Since DINEOF is not ``trained" like ML models, we cannot do conventional testing with a withheld dataset. For our experiment we use the five testing images selected in preprocessing, and then overlay artificial clouds on these images to create our testing set, which is then inserted back into the full set of images. Samples are shown in the Appendix, Figs. \ref{TestResults} and \ref{TestResultsFraser}.
After running DINEOF we then compare these reconstructions with the known full image and report accuracy.

This scheme slightly biases the experiment towards DINEOF as DINEOF has access to the cloudy testing images when generating EOFs where VConstruct does not. This is unfortunately unavoidable but the effect seems minimal.

\subsection{VConstruct Model}
The VConstruct model is based on the Variational Autoencoder \cite{Kingma2014}, it consists of an encoder, decoder and attribute network. All network layers are fully connected layers with ReLU activation functions. 

The encoder and decoder layers function exactly like they do in a conventional VAE. The encoder network compresses an image down to a lower dimensional latent space and learns a distribution it can later sample from during testing when the complete image is unknown. The 
decoder takes the output of the encoder network, or random sample from the learnt distribution, and attempts to reconstruct the original image. We use Kullback–Leibler divergence and Reconstruction loss for our loss function. 

The attribute network is based off the work of Yan et al. and Ivanov et al. \cite{Ivanov2018,Yan2016}. The network extracts an attribute vector from a cloudy image which represents what is ``known'' about the cloudy image. This attribute vector then influences the previously random image generation of the decoder network so that it generates a potential reconstruction.

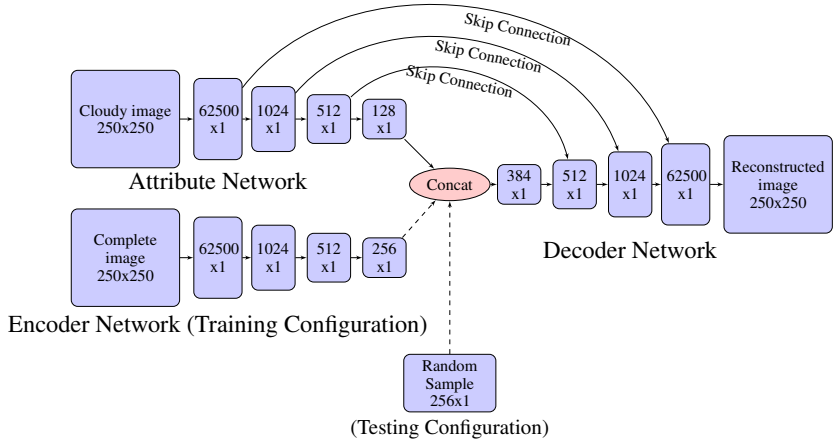
\begin{figure}
  \centering
  \adjustbox{max width=.8\textwidth}{
  \begin{tikzpicture}[node distance = 2cm, auto]
    \hspace*{-2.5cm}
    % Place nodes
    \node [block, minimum height=6em, text width=6em] (init) {Cloudy image\\250x250};
    \node [block, right of=init, minimum height=5em, text width=2.3em, node distance=2cm,label={[shift={(0,-2.6)}, scale=1.5]Attribute Network}] (flatten) {62500\\x1};
    \node [block, right of=flatten,minimum height=4em, node distance=1.2cm] (fc1) {1024\\x1};
    \node [block, right of=fc1,minimum height=3em, node distance=1.2cm] (fc2) {512\\x1};
    \node [block, right of=fc2, node distance=1.2cm] (fc3) {128\\x1};
    \node [block, below of=init, minimum height=6em, text width=6em, node distance=3cm] (Complete) {Complete image\\250x250};
    \node [block, right of=Complete, minimum height=5em, text width=2.3em, node distance=2cm,label={[shift={(0,-2.8)}, scale=1.5]Encoder Network (Training Configuration)}] (flattenC) {62500\\x1};
    \node [block, right of=flattenC,minimum height=4em, node distance=1.2cm] (fc1C) {1024\\x1};
    \node [block, right of=fc1C,minimum height=3em, node distance=1.2cm] (fc2C) {512\\x1};
    \node [block, right of=fc2C, node distance=1.2cm] (fc3C) {256\\x1};
    \node [cloud, below right of=fc3, node distance=2cm and 1.5cm] (Amalgamate) {Concat};
    \node [block, below of=Amalgamate, node distance=4.3cm,text width=5em,label={[shift={(0,-2)}, scale=1.3](Testing Configuration)}] (randomSample) {Random Sample 256x1};
    \node [block, right of=Amalgamate, node distance=1.5cm] (fc3D) {384\\x1};
    \node [block, right of=fc3D,minimum height=3em, node distance=1.2cm] (fc2D) {512\\x1};
    \node [block, right of=fc2D,minimum height=4em, node distance=1.2cm, label={[shift={(0,-2.5)}, scale=1.5]Decoder Network}] (fc1D) {1024\\x1};
    \node [block, right of=fc1D, minimum height=5em, text width=2.3em, node distance=1.2cm] (flattenD) {62500\\x1};
    \node [block, right of=flattenD, minimum height=6em, text width=6em, node distance=2cm] (Final) {Reconstructed image\\250x250};

    % Draw edges
    \path [line] (init) -- (flatten);
    \path [line] (flatten) -- (fc1);
    \path [line] (fc1) -- (fc2);
    \path [line] (fc2) -- (fc3);
    \path [line] (Complete) -- (flattenC);
    \path [line] (flattenC) -- (fc1C);
    \path [line] (fc1C) -- (fc2C);
    \path [line] (fc2C) -- (fc3C);
    \path [line] (fc3) -- (Amalgamate);
    \path [line,dashed] (fc3C) -- (Amalgamate);
    \path [line] (Amalgamate) -- (fc3D);
    \path [line] (fc3D) -- (fc2D);
    \path [line] (fc2D) -- (fc1D);
    \path [line] (fc1D) -- (flattenD);
    \path [line] (flattenD) -- (Final);
    \path [line,dashed] (randomSample) -- (Amalgamate);

    \path [line] (flatten) edge[bend left=60] node[below,pos=0.6,rotate=-12]{Skip Connection} (flattenD);
    \path [line] (fc1) edge[bend left=60] node[below,pos=0.55,rotate=-12]{Skip Connection} (fc1D);
    \path [line] (fc2) edge[bend left=60] node[below,pos=0.45,rotate=-12]{Skip Connection} (fc2D);
  \end{tikzpicture}
}
  \caption{Training Configuration for VConstruct}
  \label{training}
\end{figure}

These three networks make up the training configuration of VConstruct and can be seen in Fig. \ref{training}. When testing we cannot use the Encoder network as we do not know the complete image, so the network is replaced with a random sample from the distribution learnt in training. 
The parts that switch out are indicated by the dashed lines.

\subsection{VConstruct Testing}
We train VConstruct by using all of the complete images marked in preprocessing (minus the five testing images which are withheld) with artificial clouds overlaid. The model is trained
for 150 epochs. After training we use the five testing images, randomly selected in preprocessing, with the same artificial cloud mask as DINEOF and calculate the same metrics.

\section{Results and Discussion}
Table \ref{TestingResults} presents the results of reconstructing the five randomly selected testing days. We show results for an area off the coast of Victoria and an area by the mouth of the Fraser River.
RMSE (Root Mean Squared Error) and $R^2$ (Correlation Coefficient) are reported. The Fraser River mouth is an area of high temporal homogeneity, which is identified by Hilborn et al. as a problem area for DINEOF \cite{Hilborn2018}.
The actual reconstructed images can be found in the appendix.

\begin{table}
  \caption{Testing Results. Last row reflects overall mean performance.}
  \label{TestingResults}
  \centering
  \begin{tabular}{cccccccc}
    \toprule
    \multicolumn{4}{c}{RMSE} & \multicolumn{4}{c}{$R^2$}\\
    \cmidrule(r){1-4}
    \cmidrule(r){5-8}
    \multicolumn{2}{c}{Victoria Coast} & \multicolumn{2}{c}{Fraser River Mouth} & \multicolumn{2}{c}{Victoria Coast} & \multicolumn{2}{c}{Fraser River Mouth}\\
    \cmidrule(r){1-4}
    \cmidrule(r){5-8}
    DINEOF & VConstruct & DINEOF & VConstruct & DINEOF & VConstruct & DINEOF & VConstruct\\
    \cmidrule(r){1-4}
    \cmidrule(r){5-8}
    .104 & .125 & .183 & .152 & .247 & -.089 & .759 & .834\\
    .093 & .096 & .209 & .234 & .667 & .646  & .788 & .736\\
    .078 & .08  & .131 & .119 & .569 & .552  & .797 & .833\\
    .071 & .086 & .154 & .193 & .736 & .614  & .789 & .688\\
    .067 & .068 & .164 & .176 & .499 & .472  & .898 & .883\\
    \cmidrule(r){1-4}
    \cmidrule(r){5-8}
    .0826 & .091 & .1684 & .1748 & .544 & .439 & .806 & .791\\
    %\multicolumn{4}{c}{Averages} 
  \end{tabular}
\end{table}

For the Victoria Coast VConstruct matches DINEOF's RMSE and $R^2$ in 3/5 days but DINEOF has a better average score. For the Fraser River Mouth we see VConstruct outperforms DINEOF on 2/5 tests and nearly matches its average score, particularly in $R^2$.

\subsection{Other Benefits of VConstruct}
VConstruct also provides a few benefits unrelated to accuracy, the first being computation time. VConstruct is parallelized and runs on a GPU. 
Once trained VConstruct is able to reconstruct in roughly 10 milliseconds as opposed to the 10 minutes it took for DINEOF on the testing computer. This decrease in computation time allows researchers to reconstruct much larger datasets, which was an important concern raised by the oceanographer we consulted for this project.

VConstruct also has a few advantages that apply to DINCAE (the recent Chl-a approach from \cite{Barth2020}) as well as DINEOF. Currently VConstruct is fully atemporal, meaning that we do not need data from a previous time period to perform reconstructions. This is significant as it allows us to reconstruct data even if nothing is known about previous time periods.

Since VConstruct is based off of a VAE we can resample the random distribution to provide different possible images. From an oceanographic perspective, this allows us to generate new possible reconstructions. This is useful when subsequently collected field-truthed data was from a missing area that invalidated the initial reconstruction.
For example, the dataset we are using is field-truthed using HPLC derived Chl-a measurements from provincial ferries. 
Since reconstruction only takes a few milliseconds we could generate and test 1000s of possible images in the same time it takes for DINEOF to run.

\subsection{Future Work}
We evaluated the approach using two specific test areas. Expanding the training set by using data from other areas in the Salish Sea is important, because different oceanographic areas have different factors affecting Chl-a concentrations. The Salish Sea describes waters including Puget Sound, Strait of Georgia, and the Strait of Juan de Fuca in the US Pacific Northwest/Western Canada.
We plan on making the accuracy testing more rigorous in the next iteration.
We also plan on testing the effects of adding temporality to the input data. We initially chose to pursue atemporality as data is very commonly missing. However, temporal data is likely to improve accuracy when available.
Lastly, VConstruct uses fully connected layers for simplicity but DINCAE has shown success using convolutional layers so this will be tested in the future.

\section{Conclusion}
We have shown that VConstruct and machine learning in general can be used to reconstruct remotely sensed measurements of Chl-a, which is important in oceanographic climate change research. Even though VConstruct does not match or beat DINEOF in every accuracy test, we feel we have shown its potential for highly accurate reconstructions, particularly in areas of high homogeneity where DINEOF performs poorly. 
We also show VConstruct's other potential benefits, including better computation time as well as its ability to generate a high number of different potential reconstructions. Remote sensing is an important part of monitoring the climate and climate change, but is limited by cloud cover and other factors which
result in data loss. These factors make data reconstruction an important part of climate change research.

\section{Acknowledgements}
Special thanks to Yvonne Coady, Maycira Costa, Derek Jacoby, and Christian Marchese for their input and feedback.

\bibliography{bibliography}
\bibliographystyle{IEEEtranS}

\appendix
\section{Reconstruction Test Images}
\begin{figure}[h]
  \centering
  \includegraphics[width=\columnwidth]{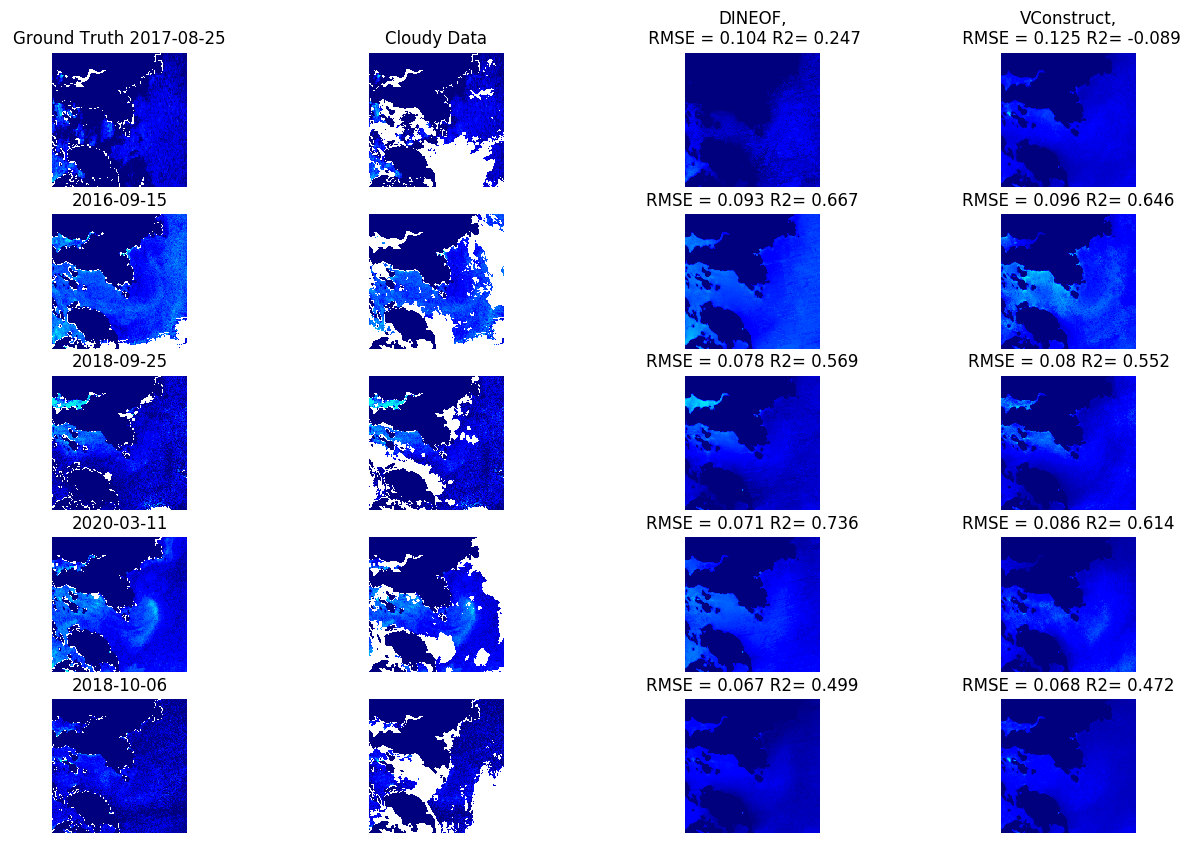}
  \caption{Test Results Victoria Coast}
  \label{TestResults}
\end{figure}
\begin{figure}
  \centering
  \includegraphics[width=\textwidth]{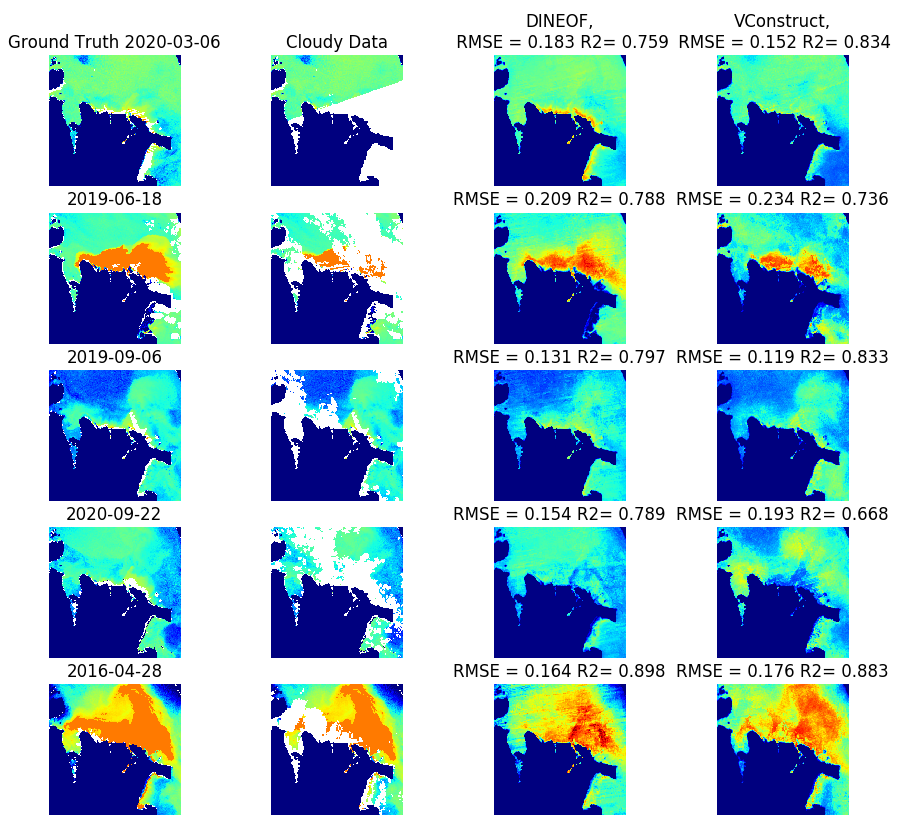}
  \caption{Test Results Fraser River Mouth}
  \label{TestResultsFraser}
\end{figure}

\end{document}